\documentclass{article}

\usepackage{microtype}
\usepackage{graphicx}
\usepackage{subfigure}
\usepackage{booktabs} 

\usepackage{hyperref}

\usepackage[accepted]{icml2019}

\usepackage[bitstream-charter]{mathdesign}
\usepackage{amsmath}
\usepackage[scaled=0.92]{PTSans}

\usepackage{wrapfig}

\usepackage[usenames,dvipsnames]{xcolor}
\definecolor{shadecolor}{gray}{0.9}

\usepackage[english]{babel}
\usepackage[parfill]{parskip}
\usepackage{afterpage}
\usepackage{framed}

{\endMakeFramed}

\usepackage{lineno}

\usepackage{ragged2e}

\newcounter{parcount}

\usepackage{graphicx}
\usepackage[labelfont=bf]{caption}

\usepackage{booktabs}

\usepackage[algoruled]{algorithm2e}
\usepackage{listings}
\usepackage{fancyvrb}
\fvset{fontsize=\normalsize}

\usepackage[all]{hypcap}
\hypersetup{citecolor=Blue}
\hypersetup{linkcolor=MidnightBlue}
\hypersetup{urlcolor=MidnightBlue}

\usepackage[nameinlink]{cleveref}

\usepackage
[acronym,smallcaps,nowarn,section,nonumberlist]{glossaries}
\glsdisablehyper{}

\lstdefinestyle{mystyle}{
    commentstyle=\color{OliveGreen},
    keywordstyle=\color{BurntOrange},
    numberstyle=\tiny\color{black!60},
    stringstyle=\color{MidnightBlue},
    basicstyle=\ttfamily,
    breakatwhitespace=false,
    breaklines=true,
    captionpos=b,
    keepspaces=true,
    numbers=left,
    numbersep=5pt,
    showspaces=false,
    showstringspaces=false,
    showtabs=false,
    tabsize=2
}
\DeclareRobustCommand{\mb}[1]{{\ensuremath{\boldsymbol{\mathbf{#1}}}}}

\newcommand{\mbt}{\mb{t}}

\newcommand{\mbz}{\mb{z}}

\newcommand{\mbbeta}{\mb{\beta}}
\newcommand{\mbdelta}{\mb{\delta}}
\newcommand{\mbepsilon}{\mb{\epsilon}}

\newcommand{\mbeta}{\mb{\eta}}

\newcommand{\mblambda}{\mb{\lambda}}

\newcommand{\mbnu}{\mb{\nu}}

\newcommand{\mbtheta}{\mb{\theta}}

\newcommand{\mbxi}{\mb{\xi}}

\newcommand{\E}{\mathbb{E}}

\newcommand{\cL}{\mathcal{L}}

\newcommand{\cN}{\mathcal{N}}

\newcommand{\g}{\,\vert\,}

\newcommand{\I}{\mathbb{I}}
\newcommand{\G}{\mathbb{G}}
\newcommand{\Hb}{\mathbb{H}}

\newtheorem{proposition}{Proposition}

\newacronym{KL}{kl}{Kullback-Leibler}
\newacronym{ELBO}{elbo}{evidence lower bound}
\newacronym{SVI}{svi}{stochastic variational inference}
\newacronym{GMM}{gmm}{Gaussian mixture model}
\newacronym{LDA}{lda}{latent Dirichlet allocation}
\newacronym{GLM}{glm}{generalized linear model}
\newacronym{MCMC}{mcmc}{Markov chain Monte Carlo}
\newacronym{BBVI}{bbvi}{black box variational inference}

\newacronym{MCLBO}{mclbo}{multiple causal lower bound}
\newacronym{MCEI}{mcei}{multiple causal estimation via information}
\newacronym{AMI}{ami}{additional mutual information}
\newacronym{MIMIC}{mimic iii}{Multiparameter Intelligent Monitoring
in Intensive Care}

\newacronym{LMM}{lmm}{linear mixed model}
\newacronym{PCA}{pca}{principal component analysis}
\newacronym{MSE}{mse}{mean squared error}
\newacronym{LOS}{los}{length of stay}

\newacronym[longplural = deep exponential families]{DEF}{def}{deep exponential family}

\newacronym{GWAS}{gwas}{genome-wide association studies}

\newacronym{RBM}{rbm}{restricted Boltzmann machine}
\newacronym{CHD}{chd}{coronary heart disease}

\newacronym{HVM}{hvm}{hierarchical variational model}
\newacronym{MIXTURE}{mixture}{}
\newacronym{NF}{nf}{normalizing flows}
\newacronym{DSVI}{dsvi}{}
\newacronym{CVI}{copula vi}{copula variational inference}

\newacronym{OPVI}{opvi}{operator variational inference}
\newacronym{LS}{ls}{Langevin-Stein}

\newacronym{CDF}{cdf}{cumulative distribution function}
\newacronym{PDF}{pdf}{probability density function}

\icmltitlerunning{Multiple Causal Inference with Latent Confounding}

\begin{document}

\twocolumn[
\icmltitle{Multiple Causal Inference with Latent Confounding}



\icmlsetsymbol{equal}{*}

\begin{icmlauthorlist}
\icmlauthor{Rajesh Ranganath}{nyu}
\icmlauthor{Adler Perotte}{columbia}
\end{icmlauthorlist}

\icmlaffiliation{nyu}{New York University}
\icmlaffiliation{columbia}{Columbia University}

\icmlcorrespondingauthor{Rajesh Ranganath}{rajeshr@cims.nyu.edu}
\icmlcorrespondingauthor{Adler Perotte}{adler.perotte@columbia.edu}

\icmlkeywords{Causal Inference}

\vskip 0.3in
]



\printAffiliationsAndNotice{}  

\begin{abstract}
Causal inference from observational data requires assumptions. These
assumptions range from measuring confounders to identifying
instruments. Traditionally, causal inference assumptions have focused
on estimation of effects for a single treatment. In this work, we construct
techniques for estimation with multiple
treatments in the presence of unobserved confounding.
We develop two assumptions based on shared confounding
between treatments and independence of treatments given the confounder.
Together, these assumptions lead to a confounder estimator regularized
by mutual information.
For this estimator, we develop a tractable lower bound.
To recover treatment effects, we use the residual information
in the treatments independent of the confounder.
We validate on simulations and an example from clinical medicine.
\end{abstract}

\section{Introduction}

Causal inference aims to estimate the effect
one variable has on another. Causal inferences form
the heart of inquiry in many domains, including
estimating the value of giving a medication to a patient,
understanding the influence of genetic variations on phenotypes,
and measuring the impact of job training programs on income.

Assumption-free causal inferences rely on randomized experimentation
\citep{cook2002experimental,pearl2009causal}.
Randomized experiments break the relationship between the intervention
variable (the treatment) and variables that could alter both the treatment 
and the outcome---confounders. Though powerful, randomized experimentation
fails to make use of large collections of non-randomized observational data
(like electronic health records in medicine) and is inapplicable where
broad experimentation is infeasible (like in human genetics). The counterpart
to experimentation is causal inference from observational data. Causal
inference from observational data requires assumptions.
These assumptions include measurement
of all confounders \citep{rosenbaum1983central}, the presence of external
randomness that partially controls treatment \citep{angrist1996identification},
and structural assumptions on the randomness \citep{hoyer2009nonlinear}.

Though causal inference from observational data has been
used in many domains, the assumptions that underlie these
inferences focus on estimation of a causal effect
with a single treatment. But many
real-world applications consist of multiple treatments.
For example, the effects of genetic variation
on various phenotypes~\citep{wellcome2007genome} or the effects of medications
from order sets in clinical medicine~\citep{o2009medical}
consist of causal problems with multiple treatments
rather than a single treatment.
Considering multiple treatments make new kinds of
assumptions possible.

We formalize \emph{multiple causal inference},
a collection of causal inference problems with multiple treatments
and a single outcome. We develop a
set of assumptions under which causal effects can be estimated
when confounders are unmeasured.
Two assumptions form the starting point:
that the treatments share confounders, and that given the shared
confounder, all of the treatments are independent. This
kind of shared confounding structure can be found in
many domains such as genetics.

One class of estimators for unobserved confounders
take in treatments and output a noisy estimate of
the unmeasured confounders. Estimators for multiple
causal inference should respect the assumptions of
the problem.
To respect shared confounding, the information between
the confounder and a treatment given the rest of the
treatments should be minimal. However, forcing this
information to zero makes the confounder independent
of the treatments. This can violate the assumption of
independence given the shared confounder.
This tension
parallels that between underfitting and overfitting.
Confounders with low information underfit,
while confounders with high information memorize
the treatments and overfit.

To resolve the tension between the two assumptions,
we develop a regularizer based on the additional
mutual information each treatment contributes
to the estimated confounder given the rest of
the treatments.
We develop an algorithm
that estimates the confounder by simultaneously minimizing the
reconstruction error of the treatments, while regularizing the additional
mutual information.
The stochastic confounder estimator can include complex
nonlinear transformations of the treatments.
In practice, we use neural networks.
The additional mutual
information is intractable, so we build a lower bound
called the \glsreset{MCLBO}\gls{MCLBO}.

The last step in building a causal estimator is to build the outcome
model. Traditional outcome models regress the confounders and
treatments to the outcome \citep{morgan2014counterfactuals}.
However, since the confounder estimate
is a stochastic function of the treatments, it contains no new
information about the response over the treatments---a regression
on both the estimated confounder and treatments can ignore the
estimated confounder. Instead, we build regression models using
the \emph{residual information} in the treatments and
develop
an estimator to compute these residuals. We call
the entire causal estimation process \gls{MCEI}.
Under technical conditions,
we show that the causal
estimates converge to the true causal effects as the number
of treatments and examples grow.
The assumptions we develop strengthen the foundation
for existing causal estimation with unobserved confounders such
as causal estimation with
\glspl{LMM}~\citep{kang2010variance,lippert2011fast}.

We demonstrate \gls{MCEI} on a large simulation study. Though traditional
methods like \gls{PCA} adjustment \citep{yu2006unified} closely approximate
the family of techniques
we describe, we find that our approach more accurately estimates the causal
effects, even when the confounder dimensionality is misspecified. Finally,
we apply the \gls{MCLBO} to control for confounders in a medical prediction problem
on health records from the \gls{MIMIC} clinical database~\citep{johnson2016mimic}. We
show the recovered effects match the literature.

\paragraph{Related Work.}

Causal inference has a long history in many disciplines
including statistics, computer science, and econometrics.
A full review is outside of the scope of this article,
however, we highlight some of recent advances in building
flexible causal models. \citet{wager2017estimation} develop
random forests to capture variability in treatment effects
\cite{wager2017estimation}. \citet{hill2011bayesian} uses
Bayesian nonparametric methods to model the outcome response.
\citet{louizos2017causal} build flexible latent variables
to correct for confounding when proxies of confounders are measured, rather
than the confounders themselves.
\citet{johansson2016learning, shalit2017estimating}
develop estimators with theoretical guarantees by building representations that
penalize differences in confounder distributions between the treated and untreated.

The above approaches focus on building estimators
for single treatments, where either the confounder or
a non-treatment proxy is measured. In contrast, computational genetics
has developed a variety of methods to control for unmeasured confounding
in \gls{GWAS}. \Acrlong{GWAS} have
multiple treatments in the form of genetic variations across multiple sites.
\citet{yu2006unified, kang2010variance,lippert2011fast}
estimate kinship matrices between individuals using a subset of the
genetic variations, then fit a \gls{LMM} where the kinship provides
the covariance for random effects. \citet{song2015testing} adjust for confounding
via factor analysis
on discrete variables and use inverse
regression to estimate individual treatment effects. \citet{tran2017implicit} build implicit
models for genome-wide association studies and describe general implicit causal models in
the same vein as \citet{kocaoglu2017causalgan}.  A formulation of multiple causal
inference was also proposed by \cite{wang2018blessings}; they take a model-based approach
in the potential outcomes framework that leverages predictive checks.

Our grounding for multiple causal inference
complements this earlier work. We develop the two assumptions
of shared confounding and of independence given shared confounders.
We introduce a mutual information based regularizer that
trades off between these assumptions.
Earlier work estimates confounders by choosing their dimensionality (e.g.,
number of PCA components)
to not overfit. This matches the flavor of the estimator we develop.
Lastly, we describe how residual information must be used to fit
complex outcome models.

\section{Multiple Causal Inference}

The trouble with causal inference from observational data lies in
confounders, variables that affect both treatments and outcome.
The problem is that the observed statistical
relationship between the treatment and outcome may be partially or completely
due to the confounder. Randomizing the treatment breaks the relationship
between the treatment and confounder, rendering the observed
statistical relationship causal. But
the lack of randomized data necessitates
assumptions to control for potential confounders.
These assumptions have focused on causal estimation
with a single treatment and a single outcome.
In real-world
settings such as in genetics and medicine, there are multiple treatments.
We now define the multiple causal inference problem, detail
assumptions for multiple causal inference,
and develop new estimators for the causal effects given
these assumptions.

Multiple causal inference consists of a collection of
causal inference problems. Consider a set
of $T$ treatments indexed by
$i$ denoted $t_i$ and an outcome $y$. The goal of
multiple causal inference is to compute
the joint casual effect of intervening on treatments
by setting them to $\mbt^*$
\begin{align*}
  \E[y \g do(\mbt = \mbt^*)]
\end{align*}
For example,
$t_i$ could be the $i$th medication for a disease
given to a patient and $y$
could be the severity of that disease. The patient's
unmeasured traits induce a relationship between the treatments
and the outcome.
The goal of multiple causal inference is to simultaneously estimate the
causal effects for all $T$
treatments. We develop two assumptions under which these
causal effects can be estimated in the presence of unobserved confounders and
later show that the estimation error gets small as the number of treatments and
observations gets large.

\paragraph{Shared Confounding.}
The first assumption we make to identify multiple causal effects is
that of
shared confounder(s). The shared confounder assumption posits
that the confounder is shared across all of the treatments.
Under this assumption, each treatment provides a view
on the shared confounder. With sufficient views, the
confounder becomes unveiled.
Shared confounding is a natural assumption in many problems.
For example, in \gls{GWAS}, treatments are genetic variations
and the outcome is a phenotype.
Due to correlations in genetic variations caused by ancestry,
the treatments share confounding.

\paragraph{Independence Given Unobserved Confounders.}
The shared confounding assumption does not identify the causal effects
since there can be direct causal links between treatments $t_i$
and $t_j$. In the presence of these links, we cannot get a clear view of the
shared confounder because the dependence between $t_i$ and $t_j$
may be due either to confounding or to the direct link between
the pair of treatments. To address this, we assume that
treatments are independent given confounders. In the discussion,
we explore strategies to loosen this assumption.

\paragraph{Implied Model.}
We developed two assumptions: shared confounding and
independence given the confounder. Together, these
assumptions imply the existence of
an unmeasured
variable $\mbz$ with some unknown distribution such that when
conditioned on, the treatments become independent:
\begin{proposition}
Let $\mbepsilon$ be independent noise terms, and $f,g,h$ be
functions. Then, shared confounding and independence given unobserved confounding imply
a generative process for the data that is
\begin{align}
\mbz &= f(\mbepsilon_\mbz), \quad t_i = h_i(\mbepsilon_i, \mbz), \quad y = g(\mbepsilon_y, \mbz, t_1, ..., t_T).
\label{eq:main-model}
\end{align}
\end{proposition}
We require that (i) any given value of the treatments, $t_i$, be
expressible as a function of the treatment noise, $\mbepsilon_i$,
given any value of the confounder, $\mbz$. Also, we require that (ii) the
outcome, $y$, be a non-degenerate function of the treatment noise,
$\mbepsilon_i$, via the treatments, $t_i$. This requirement 
ensures that there is a one-to-one mapping between $y =  g(\mbepsilon_y, \mbz, t_1, ..., t_T)$ and the rewritten version $y =  g(\mbepsilon_y, \mbz, h_1(\mbepsilon_1, \mbz), ..., h_T(\mbepsilon_T, \mbz))$.

If the model in \Cref{eq:main-model} were explicitly given, posterior inference
would reveal the
causal effects up to the posterior variance of the latent confounder.
In general the model is not known, thus the goal is to build
a flexible estimator for a broad class of problems.

\section{Unobserved Confounder Estimation in Multiple Causal Inference}

We develop an estimator for the unobserved confounder in multiple causal
inference without directly specifying a generative model.
This estimator finds the confounder
that reconstructs each treatment given the other treatments. The
estimator works via information-based
regularization and cross-validation in a way that is agnostic to
the particular functional form of the estimator.
We present a pair of lower bounds to estimate the confounder.

\paragraph{Confounder Estimation.}
The most general form of a confounder estimator
is a function that takes the following as input:
noise $\mbepsilon$, parameters $\mbtheta$, treatments
$\mbt_j$, and outcome $y_j$. Using the outcome without extra assumptions
is inherently ambiguous. The ambiguity lies in
how much of $y_j$ is retained in $\mbz_j$. The only statistics we observe about
$y$ come from $y$ or its cross statistics with $\mbt$. From \cref{eq:main-model},
we know that the cross statistics provide a way to estimate $\mbz$
the confounder. However, since the outcome depends on the treatments, cross
statistics between the treatment and outcome could either be from the
confounder or from the direct relationship between the treatments and outcome.
This ambiguity cannot be resolved without further assumptions like assuming a model.
Therefore we focus on estimating the unobserved confounder without using the
outcome, where the outcome has been marginalized.

A generic stochastic confounder estimator with marginalized outcome
is a stochastic function of the treatments controlled by parameters
$\mbtheta$. The posterior of the latent confounder in a
model is an example of such a stochastic function.
To respect the assumptions, we want to find a
$\mbtheta$ such that
conditional on the confounder, the treatments are independent.
The trivial answer to this estimation problem is to have
the confounder memorize the treatments.
We develop a regularizer based on the information the confounder
retains about each treatment.

\paragraph{Additional Mutual Information.}
We formalize the notion of information using mutual information~\citep{cover2012elements}.
Let $\I(a,b)$ denote the mutual information.
Mutual information is nonnegative and is zero when $a$ and $b$ are independent.
To understand the flexibility in building stochastic confounder estimators,
consider the information between the estimated confounder and treatment $i$ given the remaining
treatments $\mbt_{-i}$: $\I(t_i, \mbz \g  \mbt_{-i})$. We call this the
\gls{AMI}. It is the additional information a treatment can provide
to the confounder, over what the rest of the treatments provide.
The \acrlong{AMI} takes values
between zero and some nonnegative number. The maximum indicates
that $\mbz$ and $\mbt_{-i}$ perfectly
predict $t_i$. When all variables are discrete, the upper bound
is the entropy $\Hb(t_i \g \mbt_{-i})$. This range parameterizes
the flexibility in how much information the confounder encodes
about treatment $i$, over the information present in the remaining treatments.

At first glance, letting $\I(t_i, \mbz \g  \mbt_{-i}) > 0$ seems to
violate shared confounding because the confounder $\mbz$ has information
about a treatment that is not in the other treatments. But setting
$\I(t_i, \mbz \g  \mbt_{-i}) = 0$ forces the confounder to be independent
of all of the treatments. The shared confounder assumption is
in tension with the assumption of the independence of
treatments given the confounder. Since if the confounder-estimated
entropy $\Hb(t_i \g \mbt_{-i})$ is bigger
than the true entropy under the population sampling distribution $F$, $\Hb_F(t_i \g \mbt_{-i})$, the treatments cannot be independent given the confounder.

From the perspective of confounder estimation, the two
assumptions can be seen as underfitting and overfitting.
Satisfying the shared confounding assumption leads to
underfitting since no information goes to the confounder.
While independence of treatments given confounder favors
overfitting by pushing all treatment information into the
confounder.

\paragraph{Regularized Confounder Estimation.}

The estimator controls the \gls{AMI}~~$\I(t_i, \mbz \g  \mbt_
{-i})$ via regularization.
For compactness, we drop the unobserved confounder estimator's functional
dependence on $\mbepsilon$ and
write $\mbz \sim p_\mbtheta(\mbz \g \mbt)$.
Let $p(\mbt)$ be the empirical distribution over the observed treatments, let
$\mbbeta$ be a parameter, and let $\alpha$ be a regularization parameter.
Then we define an objective that
tries to reconstruct each treatment independently given $\mbz$, while controlling
the \acrlong{AMI}:
\begin{align}
\max_{\mbtheta, \mbbeta} \quad \E_{\mbt \sim p(\mbt)} & \E_{p_\mbtheta(\mbz \g \mbt)} \left[\sum_{i=1}^T \log p_{\mbbeta}(t_i \g \mbz) \right]
\nonumber \\
- & \alpha \sum_{i=1}^T \I_\mbtheta(t_i, \mbz \g  \mbt_{-i}).
\label{eq:mc}
\end{align}
We will suppress the index $i$ in $p_\mbbeta$ when clear.
The distributions $p_\mbbeta$ and $p_\mbtheta$ can be from any class
with tractable log probabilities;
in practice we use conditional distributions built from neural networks,
e.g., $\mbz \sim \textrm{Normal}(\mu_\mbtheta(\mbt), \sigma_\mbtheta(\mbt))$, where
$\mu$ and $\sigma$ are neural networks.
This objective finds the $\mbz$ that can reconstruct $\mbt$ most
accurately, assuming the treatments are conditionally independent given $\mbz$.
\Cref{eq:mc} can be viewed as an autoencoder where the code is regularized
to limit the \acrlong{AMI}, thereby preferring to keep information this is
shared between treatments.

The information regularizer is similar to regularizers in supervised learning.
Consider how well the confounder predicts a treatment when
estimated conditional on the rest of the treatments. When $\alpha$ is too small for a
flexible model, the confounder
memorizes the treatment so the prediction error $\Hb(t_i \g \mbt_{-i})$ is big.
When $\alpha$ is too large, $\mbz$ is independent of the treatments so again the prediction error is big.
This mirrors choosing the regularization
coefficient in linear regression. When the regularization is too large, the regression coefficients
ignore the data, and when it is too small, the regression coefficients memorize the data.
As in regression, $\alpha$ can be found by cross-validation. Minimizing
the conditional entropy directly rather than by cross-validation leads to the degenerate
solution of $\mbz$ having all the information in $\mbt$.

Directly controlling the additional mutual information contrasts classical
latent variable models,
where tuning parameters like the dimensionality,
flexibility of the likelihood, and number of
layers in a neural network implicitly controls the additional mutual information.

Since we do not have access to $p(t_i, \mbz \g \mbt_{-i})$, the objective
contains an intractable mutual information term.
We develop a tractable objective based on lower bounds
of the negative mutual information.

We develop two lower bounds for the negative \gls{AMI}. The
first bounds the entropy, while the second introduces an auxiliary
distribution to help make a tight bound.

\paragraph{Direct Entropy Bound.}
The conditional mutual information can be written in terms of conditional entropies
as
\begin{align*}
 \I_\mbtheta(t_i, \mbz \g  \mbt_{-i}) &= \Hb_\mbtheta(\mbz \g \mbt_{-i}) - \Hb_\mbtheta(\mbz \g \mbt_{-i}, \mbt_i)
\\
 &= \Hb_\mbtheta(\mbz \g \mbt_{-i}) - \Hb_\mbtheta(\mbz \g \mbt).
\end{align*}
The second term comes from the entropy of $p_\mbtheta(\mbz \g \mbt)$ and is tractable
when the distribution of the confounder estimate is known.
But
the first term requires marginalizing out the treatment $t_i$. This conditional
entropy with marginalized treatment is not tractable, so we develop
a lower bound. Let $p(\hat{t}_i)$ be the marginal distribution of treatment $i$
and C be a constant with respect to $\mbtheta$;
expanding the integral gives
\begin{align*}
-\Hb_\mbtheta(\mbz \g \mbt_{-i})
&\geq \int p(\mbz \g \mbt) p(\mbt) p(\hat{t}_i) \log p(\mbz \g
 \mbt_{-i}, \hat{t}_i) d\mbz d\mbt + C
\end{align*}
The lower bound follows from Jensen's inequality. A full derivation
is in the appendix.
Unbiased estimates of the lower bound can be computed via Monte Carlo.
Substituting this back, the information-regularized confounder estimator objective gives
a tractable lower bound to the information-regularized objective.

\paragraph{Lower Bound via Auxillary Distributions.}
The gap between the information regularizer and
the direct entropy lower bound may be large. Here,
we introduce a lower bound with parametric
auxillary distributions whose parameters can be optimized
to tighten the bound. Let $r$ be a probability distribution
with parameters $\mbxi_i$, then a lower bound on the
negative mutual information is
\begin{align*}
-\I(t_i;\mbz \g \mbt_{-i}) &= -\textrm{KL}(p(t_i;\mbz \g \mbt_{-i}) || p(t_i \g
\mbt_{-i}) p(\mbz \g \mbt_{-i}))
\\
&\geq -\E_{p(\mbt) p( \mbz \g \mbt)} \log \frac{p(\mbz \g \mbt)}{r_{\mbxi_i}(\mbz \g \mbt_
{-i})}
\\
&:= \G_{\mbtheta, {\mbxi_i}}(\mbz \g \mbt_{-i}).
\end{align*}
This bound becomes tight when $r(\mbz \g \mbt_
{-i} ; \mbxi_i)$ equals $p(\mbz \g \mbt_
{-i})$, the condition under the confounder estimator.
We derive this lower bound in detail in the appendix.
Substituting this bound into the information-regularized
confounder objective gives
\begin{align}
\cL =& \E_{\mbt \sim p(\mbt)} \E_{p_\mbtheta(\mbz \g \mbt)} \left[\sum_{i=1}^T \log p_\mbbeta(t_i \g \mbz) \right]
+ \alpha \sum_{i=1}^T \G_{\mbtheta, {\mbxi_i}}(\mbz \g \mbt_{-i}).
\label{eq:mc-lbo}
\end{align}
We call this lower bound the \glsreset{MCLBO}\gls{MCLBO}.

\paragraph{Algorithm.}
To optimize the \gls{MCLBO}, we use stochastic gradients by passing the derivative
inside expectations \citep{williams1992simple}. These techniques underly black
box variational inference algorithms \citep{ranganath2014black, Kingma:2014, rezende2014stochastic}.
We derive the full gradients for $\mbbeta$, $\mbtheta$, and $\mbxi$ in the
appendix. With these
gradients, the algorithm can be summarized in \Cref{alg:confound-lower-bound}. We choose a range of
$\alpha$ values
and fit the confounder estimator using the \gls{MCLBO}. We then select the $\alpha$ that minimizes the
entropy $\sum_i \Hb(\mbt_i \g \mbt_{-i})$ on held-out treatments. In practice, we allow a small
relative tolerance for larger $\alpha$'s over the best held-out prediction to account for finite sample estimation
error. The algorithm can be seen as learning an autoencoder.
The code of the this autoencoder minimizes the information retained about each treatment
subject to the code predicting each $t_i$ best when the code is computed only from $\mbt_{-i}$,
the remaining treatments.

\begin{algorithm}[t]
\SetKwInOut{Input}{Input}
\SetKwInOut{Output}{Output}
 \Input{Reconstruction: $p_\mbbeta(\mbt_i \g \mbz)$, \newline Stochastic Confounder Estimate
 $p_\mbtheta(\mbz \g \mbt)$, \newline Lower Bound $r_{\mbxi_i}(\mbz \g \mbt_
 {-i})$ \newline Vector of $\alpha$}
 \Output{Confounder Estimate Parameters: $\mbtheta$}
 Initialize $\mbbeta$ and $\mbtheta$ randomly.

\For{each $\alpha$} {
  \While{not converged}{
    Compute unbiased estimate of $\nabla_{\mbtheta} \cL$.
    \quad (\cref{eq:theta-gradient}) \\
   Compute unbiased estimate of $\nabla_{\mbbeta} \cL$. \quad
    (\cref{eq:beta-gradient}) \\
    Compute unbiased estimate of $\nabla_{\mbxi} \cL$. \quad
    (\cref{eq:xi-gradient}) \\
    Update $\mbtheta$, $\mbbeta$, $\mbxi$ using stochastic gradient ascent. \\
  }
 }
 Return $\mbtheta$ for best $\alpha$ in $\sum \Hb(\mbt_i \g \mbt_{-i})$ on held
 out
 data.
 \caption{Confounder estimation via lower bound}
 \label{alg:confound-lower-bound}
\end{algorithm}

\section{Estimating the Outcome Model}
In traditional observational causal inference, the possible outcomes are independent of the treatments given the confounders, so predictions given confounders are
causal estimates. With the $do$ notation that removes
any influence from confounding variables, we have
\begin{align*}
\E[y \g do(\mbt = \mbt^*)] &= \E_{p(\mbz)}\E[y \g do(\mbt = \mbt^*), \mbz]
\\
&= \E_{p(\mbz)} \E[y \g \mbt^*, \mbz].
\end{align*}
So to estimate the causal effect, it suffices to build a consistent regression
model. However with estimations of unobserved confounder that
are stochastic functions of the treatment, this relationship
breaks:
$\I(\mbz, y) \leq \I(\mbt, y)$ and $\I(y, \mbz \g \mbt) = 0$.
The confounder has less information
about the outcomes than the treatments themselves. Given
the treatments, the confounders provide no information about
the outcome.
The lack of added information means that if we were to simply
regress $\mbt$ and  $\mbz$ to $y$, the regression could completely ignore the
confounder. Building outcome models for each $\mbz$ addresses
this issue, but requires doing separate regressions for every possible
$\mbz$.

A regression conditional on the confounder can only
treatment variation independent of the confounder.
Recovering these
independent components makes outcome estimation feasible.
Formally, let
$\mbepsilon_i$ be the independent component of the $i$th treatment,
then we would like to find a distribution $ p(\mbepsilon_i \g \mbz, \mbt)$
that maximizes
\begin{align}
\E_{p(\mbt) p_\mbtheta(\mbz \g \mbt) \prod_i p(\mbepsilon_i \g \mbz, t_i)} \left
[\sum_{i=1}^T \log p(t_i \g \mbz, \mbepsilon_i) \right],
\nonumber \\
 \textrm{such that } \I(\mbepsilon_i, \mbz) = 0.
\label{eq:epsilon-estimation}
\end{align}
Optimizing this objective over $p(\mbepsilon_i \g \mbz, \mbt)$ and $p(t_i \g \mbz, \mbepsilon_i)$
provides stochastic estimates of the part of $t_i$ independent of $\mbz$. We call this leftover part $\mbepsilon_i$ the residuals.
These residuals are independent of the confounders and therefore $\I(y, \mbz \g
\mbepsilon) = \I(y, \mbz)$, so when regressing confounders and residuals
to the outcome, the confounding is no longer ignored.
The residuals are a type of instrumental variables; they are
independent and affect
the outcome only through the treatments.

Optimizing \cref{eq:epsilon-estimation} can be a challenge both
due to the intractable mutual information constraint and that $p(t_i \g \mbz, \mbepsilon_i)$
may have degenerate density.
In the appendix, we provide a general estimation technique for the residuals. Here we focus on
a simple case where $t_i \sim p_{\mbbeta}(t_i \g \mbz)$ in
\cref{eq:mc} can be for, some function $d$,
written as $t_i = d(\mbz, \mbepsilon_i)$ for $\mbepsilon_i$ drawn independently. Then if $d$ is invertible for every fixed value of $\mbz$, the residuals $\mbepsilon_i$ that satisfy \cref{eq:epsilon-estimation} can be found
via inversion. That is, \cref{eq:epsilon-estimation} is optimal if $\mbepsilon_i = d^{-1}(\mbz, t_i)$, since
$\mbepsilon_i$ is independent of $\mbz$ by construction and
in conjunction with $\mbz$ perfectly reconstructs $\mbt_i$.
This means when the reconstruction in \cref{eq:mc} is invertible,
independent residuals are easy to compute.

To learn the outcome model, we regress with the residuals and confounder by maximizing
\begin{align}
\max_{\mbeta} \,\, \E_{p(y, \mbt) p_\mbtheta(\mbz \g \mbt) p(\mbepsilon_i \g
\mbz,
\mbt)} [\log
p_{\mbeta}(y \g \mbz, \mbepsilon)].
\label{eq:outcome}
\end{align}
Since $\mbepsilon$ and $\mbz$ are independent, they provide different information to $y$.
To compute the causal estimate, $p(y \g \mbz, do(\mbt))$ given $p(y \g \mbz, \mbepsilon)$,
we can substitute:
\begin{align}
p(y \g \mbz, do(\mbt = \mbt^*)) = p(y \g \mbz, \mbt^*) = p(y \g \mbz, \mbepsilon = d^{-1}(\mbt^*, \mbz)).
\label{eq:outcome-do}
\end{align}
The right hand side is in terms of known quantities: the outcome model from \cref{eq:outcome} and the $\mbepsilon$
from the confounder estimation in \cref{eq:mc-lbo}. The causal estimate of $do(\mbt = \mbt^*)$
can be computed by averaging over $p(\mbz)$. We call the process of confounder estimation
with the \gls{MCLBO} followed by outcome estimation with residuals,
\glsreset{MCEI}\gls{MCEI}.

\paragraph{Casual Recovery.}
It is not possible to recover latent variables without strong
assumptions such as a full probabilistic model.
Any information preserving transformations
of the confounder estimate look equivalent to the \gls{MCLBO}.
However this is not a problem.
The outcome model and
estimation with \Cref{eq:outcome-do}
produce the same result for information preserving transformations.

Recovering the residual for each treatment and the latent confounder
from
the treatments requires finding $T + 1$ variables from $T$ variables;
this has many possible solutions.
The existence of an extra variable might raise concerns
about the ability to find a solution or that positivity, all treatments will
occur for each value of the confounder, is met. Both of these issues
become smaller as the number of treatments grows.
We formalize causal recovery with \gls{MCEI} in a simple setting.
\begin{proposition}
  If the confounder is finite dimensional and the treatments are i.i.d.
  given the confounder,
  then the multiple causal estimator in \cref{eq:mc} combined with \cref{eq:outcome-do}
  recovers the correct causal estimate as $T \to \infty$, $N \to \infty$.
\end{proposition}
The intuition is that as $T \to \infty$, we get perfect
estimates of $\mbz$ up to information equivalences. The amount of
information about each treatment in the confounder given the
rest of the treatments
goes to zero, so the assumption of shared confounding and independence are
both satisified.
Asymptotics require constraints for the outcome model to be well defined.
For example, with normally distributed outcomes the variance needs to be finite~\citep{AmourBlog2018}.
This proposition generalizes to non-identically distributed treatments where
posterior concentration occurs.

Positivity in this proposition gets satisfied by the fact that we
required earlier that (i) $y$, be a non-degenerate function of the treatment
noise meaning that $y$ can only be written as a function of the confounder and treatment in one way even asymptotically and
(ii) any given value of the treatments, $t_i$, be
expressible as a function of the treatment noise, $\mbepsilon_i$,
given any value of the confounder,
and that the confounder estimate converges to a constant for each data point.
To see this constructively, let $\mbt_{e}$ be the treatments
with even index and let $\mbt_{o}$ be the treatments with odd index.
Recall that if $I(\mbt_{o} ; \mbz \g \mbt_{e}) = I(\mbt_{e} ; \mbz \g \mbt_
{o}) = 0$, then the pair $\mbt_{o}, \mbt_{e}$ is independent of $\mbz$.
As $T$ gets large, both additional mutual informations, $I(\mbt_{o} ; \mbz \g
\mbt_{e}), I(\mbt_{e} ; \mbz \g \mbt_{o})$, tend to zero. The
estimated confounder becomes independent. The assumption (i) of
non-degenerate response functions of the treatment noise rules out outcome models
that asymptotically only depend on the confounder, while assumption (ii)
rules out treatment values reachable by only certain unmeasured confounder
values. Together, along with the estimated $\mbz$'s asymptotic independence
of the treatments ensure that positivity is met.

\begin{figure*}
\centering
\begin{subfigure}{}
            \centering
  \includegraphics[width = .48\linewidth]{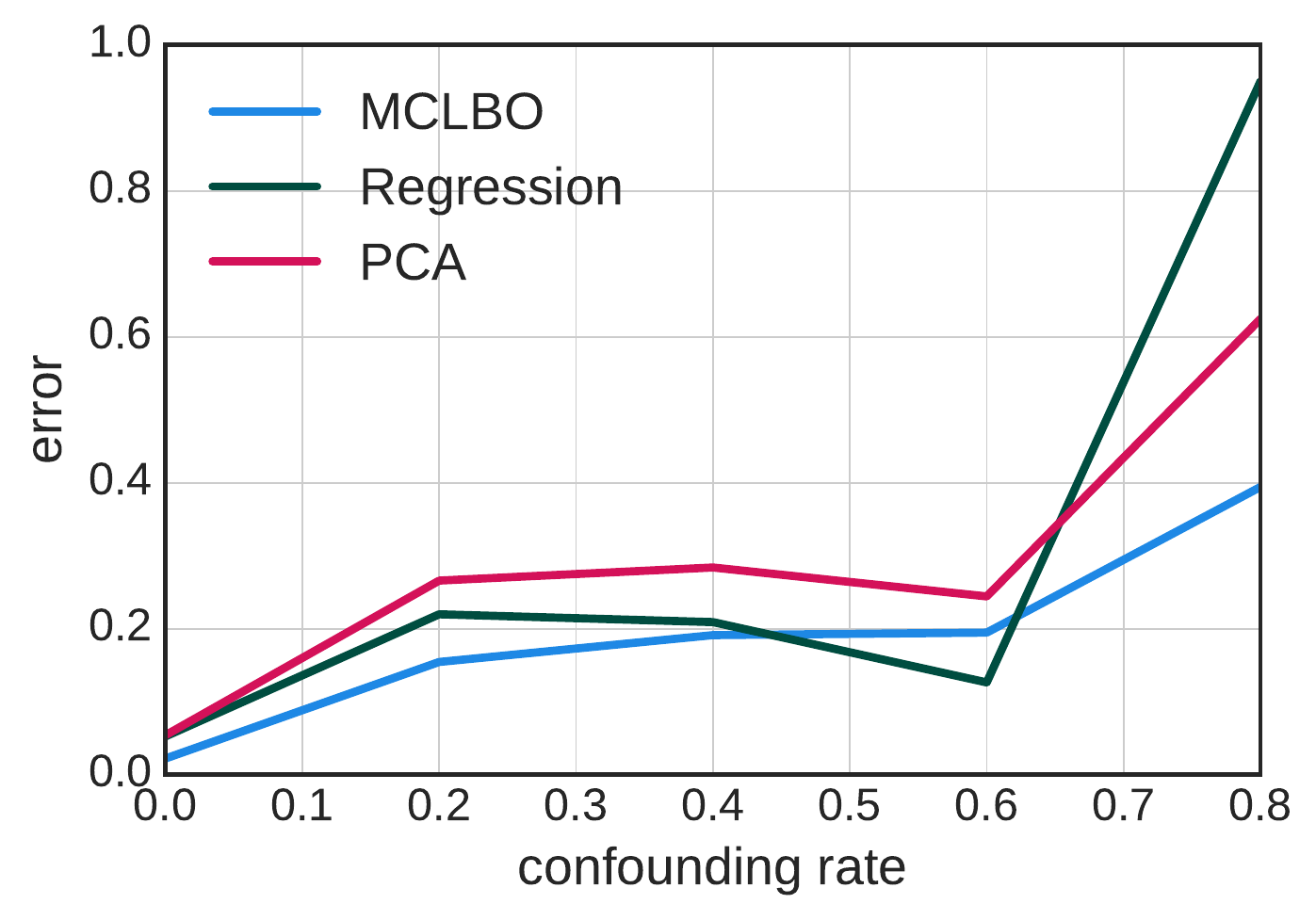}
 \end{subfigure}
 \begin{subfigure}{}
             \centering
   \includegraphics[width = .48\linewidth]{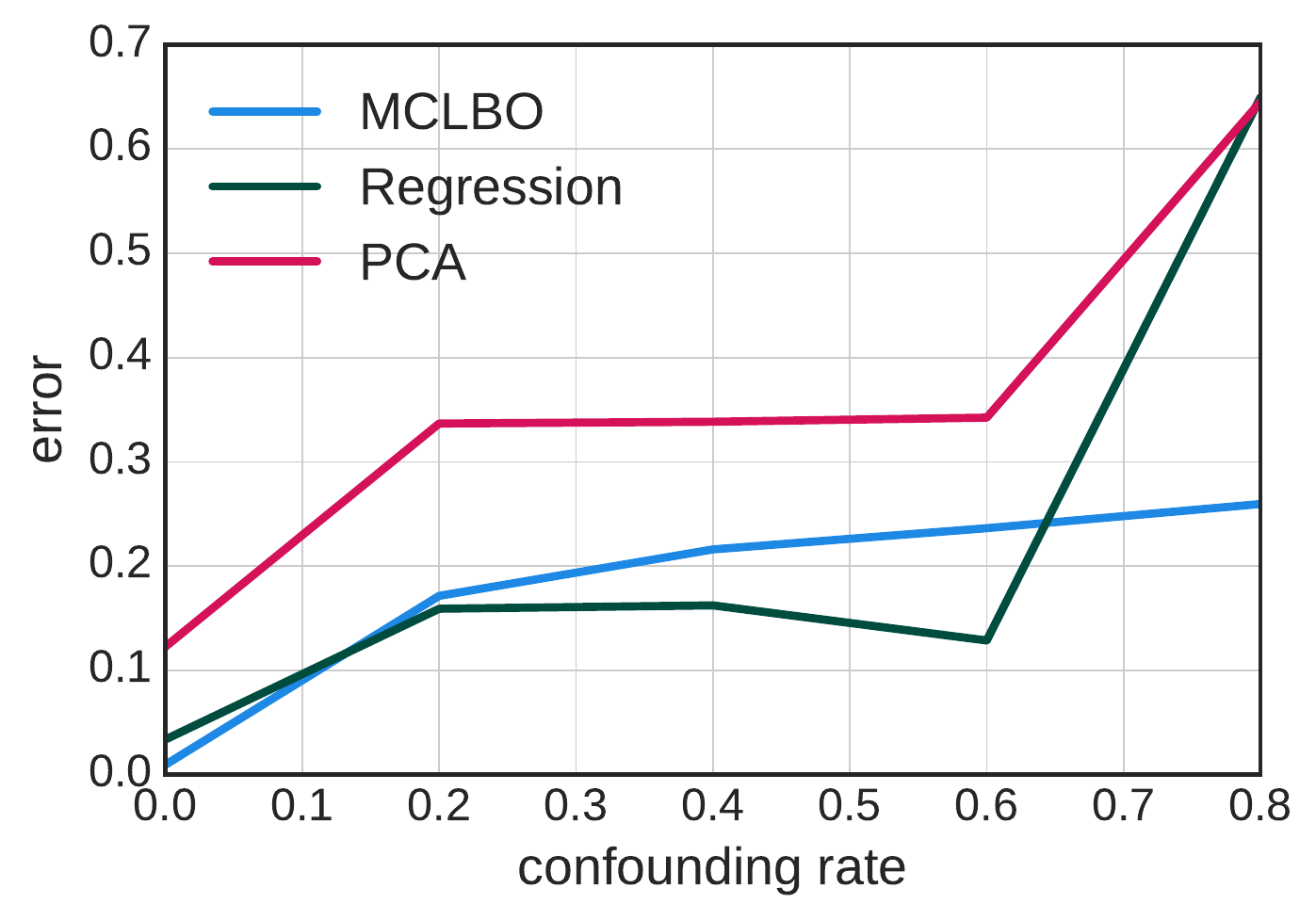}
 \end{subfigure}
 \vskip -.15in
\caption{Simulation results for the correctly specified (left) and overspecified
(right) confounder dimensionality. \gls{MCEI} performs similar or better than
\gls{PCA} and better
 tolerates misspecification. Basic regression performs poorly at high
 levels of confounding.}
  \label{fig:simulation}
\vskip -.2in
\end{figure*}

\paragraph{Flexible Estimators.}
With flexible estimators, there can be an issue that
leads to high estimator variance. We illustrate this
issue with the simple case of independent treatments.
Suppose the true treatments $\mbt_j$ come from an unconfounded model, where
all of the $\mbt_j$ are independent.
Consider using a latent variable model for confounder estimation,
where each observation has
a latent variable $\mbz$ and treatment vector $\mbt$.
Let $W$ be a matrix of parameters, let $\kappa$ and $\sigma$
be hyperparameters, and let $j$ index observations.
Then the model is $
\mbz_j \sim \cN(0, \sigma), \quad \mbt_j \sim \cN(W \mbz_j, \kappa)$.
The maximum likelihood estimate for the this model with latent size equal to
data size
given this true model
is $W^* = I (1 - \kappa)$, up to rotations. Posterior distributions
are a type of stochastic confounder estimator. Here, the posterior distribution
is
\begin{align*}
p(\mbz \g \mbt) = \cN\left(\frac{\sigma}{\sigma + \kappa} \mbt, \sigma^2 \left(1 - \frac{\sigma}{\sigma + \kappa}\right)\right).
\end{align*}
A range of estimators given by this posterior indexed by $\kappa$
have the same
conditional entropy and predictive likelihoods.
All residuals for any $\kappa > 0$ recover the right causal effect.
The trouble comes in as $\kappa$ gets small. Here the variance in the residual
gets small, which in turn implies the variance of the
estimated causal effect gets large similar to when using a weak instrument.
In the appendix, we provide families of estimators that exhibit this
problem for general treatments. If the estimator family is rich
enough there may be multiple, \gls{AMI} regularization
values $\alpha$ that lead to the same prediction.
Choosing the largest value of \gls{AMI} regularization mitigates this issue by
find the estimator in the equivalence class that leaves the most information
for the residuals, thus reducing the variance of the causal estimates.

\section{Experiments}

We demonstrate our approach on a large simulation
where the noise also grows with the amount of confounding.
We study variants of the simulation where the estimators
are misspecified.
We also study a real-world example from
medicine. Here, we look at the effects of various lab
values prior to entering the intensive care unit on the length
of stay in the hospital.

\paragraph{Simulation.}
Consider a model with real-valued treatments.
Let $n$ index the observations and $i$ the treatments.
Let $W$ be a parameter matrix, $\sigma$ be the simulation
standard deviation, $\gamma$ be the confounding rate,
and $D$ be the dimensionality
of $\mbz$.
The treatments are drawn conditional on
an unobserved $\mbz_n$ as
\begin{align}
\mbz_n \sim \textrm{Normal}&(0, \gamma), \quad \mbepsilon_n \sim \textrm{Normal}
(0, 1 - \gamma), \nonumber \\
&t_{i, n} \sim \textrm{Normal}(W \mbz + \mbepsilon_n, \sigma),
\label{eq:treatment-gen}
\end{align}
where $\gamma$ scales the influence of $\mbz_n$ on each of
the treatments. Let $b$ be weight vectors and
$\sigma_y$ be the outcome standard deviation. Then the outcomes are
\begin{align*}
y_n \sim \textrm{Normal}((1 - \gamma) b_{\epsilon}^\top \mbepsilon_n +  \gamma
b_{\mbz}^\top
|\mbz_n|, \sigma_y).
\end{align*}
The amount of confounding grows with $\gamma$. The simulation is nonlinear
and as the confounding rate grows, the estimation problem becomes harder
because the $\sigma_y$ gets larger relative to the effects.

We compare our approach to the \gls{PCA} correction
\citep{yu2006unified} which closely relates to \gls{LMM}
\citep{kang2010variance,lippert2011fast}.
These approaches should perform well in confounding
estimation since the process
in \cref{eq:treatment-gen} matches the assumptions behind probabilistic
\gls{PCA}~\citep{christopher2016pattern}. However because of the outcome
information problem from the previous section,
the results may be vary depending on whether the outcome model uses the
estimated confounders.
For confounder estimation by the \gls{MCLBO},
we limit \gls{MCEI} to have a similar number of parameters. Details
are in the appendix.

We study two cases. First, we correctly give all methods the right
dimensionality $D=2$. Second, we misspecify: all methods use a bigger dimension
$10$, while the true $D=2$.
We measure MSE to the true parameters scaled by the true norm. We simulate
$10,000$ observations with $50$ treatments over $5$ redraws. We use neural
networks for all functions and Gaussian likelihoods. We describe the
remaining simulation parameters in the appendix.

\Cref{fig:simulation} shows the results. The left panel plots the error for varying levels of confounding when the confounder dimension is correctly specified. We find that confounder
estimation with \gls{MCEI} performs similar to or better than \gls{PCA}.
Regression performs poorly as the confounding grows. Though \gls{PCA}
is the correct model to recover the unobserved confounders, the outcome model
can ignore the confounder due to the information inequality in the previous
section. The right panels show \gls{MCEI} tolerates
misspecification better than \gls{PCA}.

\paragraph{Clinical Experiment.}

\begin{figure}[!t]
  \begin{center}
    \includegraphics[width=0.48\textwidth]{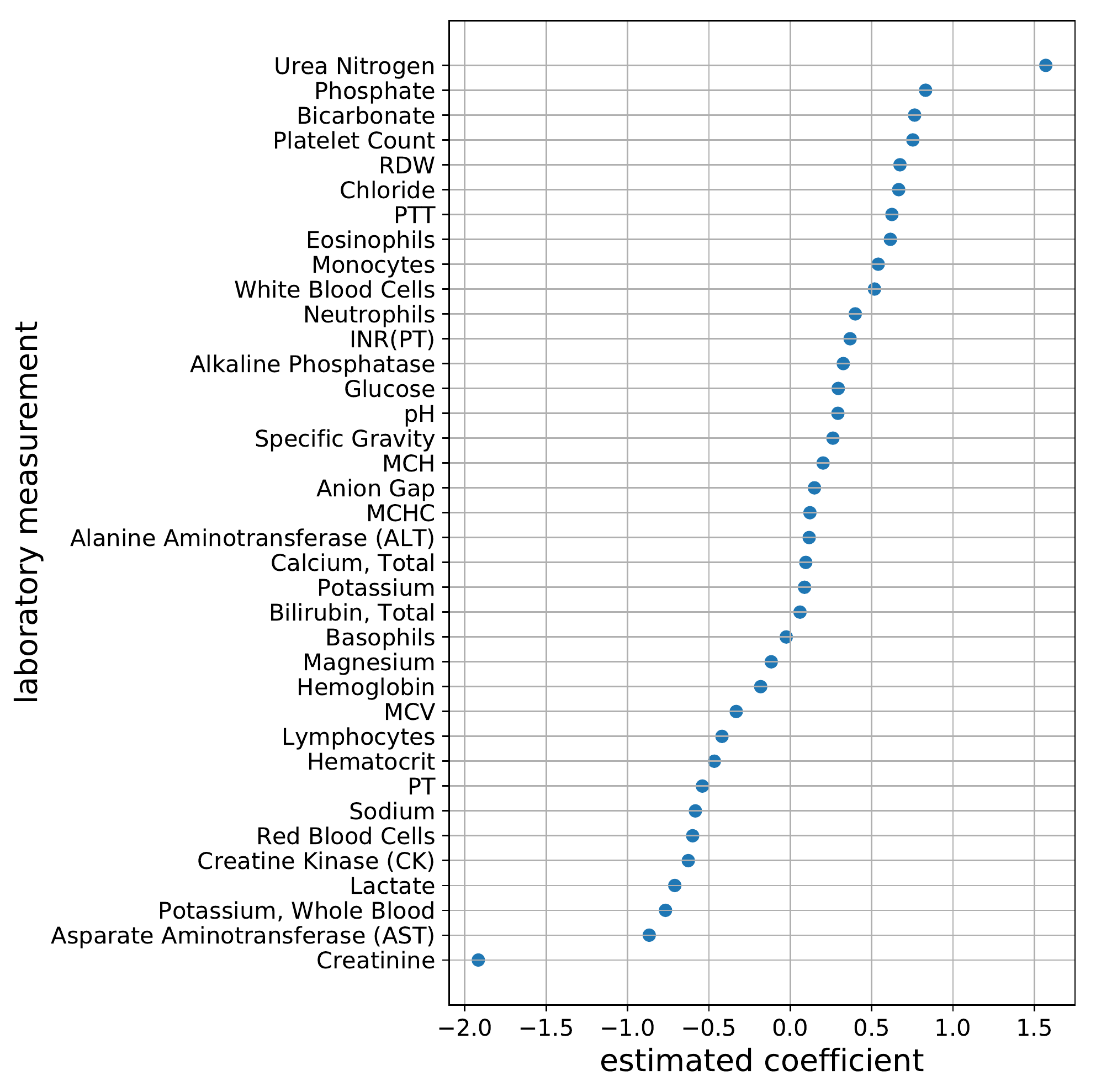}
  \end{center}
  \vskip -.15in
  \caption{Causal estimates for effects of laboratory values on ICU length of stay.}
  \label{fig:experiment}
  \vskip -.2in
\end{figure}

\Gls{LOS} is defined as the duration of a hospital visit.
This measure is often used as an intermediate outcome in studies due to the
associated adverse primary outcomes. Patient flow is important
medically because unnecessarily prolonged hospitalization places
patients at risk for hospital acquired infections (among other adverse
outcomes). These can be difficult to treat and are associated with
significant morbidity, mortality, and cost. Studies have found a
1.37\% increase in infection risk and 6\% increase in any adverse
event risk for each excess \gls{LOS} day~\citep{hassan2010hospital,
  andrews1997alternative}. Also, it is of operational concern for
hospitals because reimbursement for medical care is increasingly tied
to visit episodes rather than to discrete products or services
provided~\citep{press2016medicare}.

The dataset studied in this experiment is comprised of 25753 ICU visits
and 37 laboratory tests from the \gls{MIMIC} clinical database~\citep{johnson2016mimic}. We
applied our \gls{MCEI} approach to laboratory tests measured in the
emergency department prior to admission as treatments, and a binarized
\gls{LOS} based on the average value as outcome. Laboratory test values were
shifted, log-transformed, and standardized with missing values
imputed to the median of the laboratory test type.

The results are shown in \cref{fig:experiment} and correlate well with
findings in the literature regarding factors influencing \gls{LOS}.
For example, elevated blood urea nitrogen is associated with states of
hypovolemia (prerenal azotemia) and hypercatabolism and has been
linked to increased \gls{LOS} in pancreatitis and stroke
patients~\citep{faisst2010elevated, lin2015preliminary}. Elevated
white blood cells, or leukocytosis, is one of the main markers for
infection and, as expected, infection has been associated with
increased \gls{LOS}, particularly when
systemic~\citep{beyersmann2009nosocomial,talmor1999relationship}.
Other findings, such as an inverse relationship to potassium
(hypokalemia) is also supported by the
literature~\citep{paltiel2001management}. An inverse relationship with
creatinine may be due to age, a confounder that was not included and
likely violates the shared confounding because it primarily only
effects creatinine.

\section{Conclusion}

We formalized two assumptions needed for multiple causal inference, namely shared
confounding and independence between treatments given the shared confounders.
Together, these assumptions imply a information-regularized estimator
for unmeasured confounders.
We developed lower bounds for a tractable algorithm.
We showed how stochastic residuals can be used to estimate the outcome model, and
we demonstrated our approach in simulations and on ICU data.

Many future directions
remain. First, the assumptions we made are likely not tight. For example, the independence
between treatments given the shared confounder could be relaxed to allow
a finite number of dependencies between observations. The intuition is that if there is
a limited amount of dependence between treatments, the confounder can be estimated from
the other treatments. Next, in the algorithm to estimate the information, the lower bound can
be replaced by likelihood ratio estimation. This has the benefit of removing slack
in the bound, while also improving numerical stability by avoiding
differences. Finally, with multiple
outcomes, new kinds of estimators that are simpler can be developed.

\subsection*{Acknowledgments}

We would like to acknowledge Jaan Altosaar, Fredrik Johansson, Rahul Krishnan, Aahlad Manas Puli,  and Bharat Srikishan for helpful discussion and comments.

\bibliographystyle{apalike}
{
\bibliography{multicausal}
}

\newpage

\appendix
\onecolumn
\section{Appendix}
\paragraph{Big Estimator Classes.}
Stochastic confounder estimators can be
constructed by looking at posteriors of models.
We construct a big estimator class by building
two models that match the observed data and looking
at mixtures of these two models.
Both models have the same distribution of treatments and
outcomes and have treatments that are independent of the
outcome. Take the model
\begin{align*}
\mbz &= f(\mbepsilon_\mbz) \nonumber \\
t_i &= h_i(\mbepsilon_i, \mbz) \nonumber \\
y &= g(\mbepsilon_y, \mbz, t_1, .... t_T),
\end{align*}
and the model
\begin{align*}
\mbz &= f(\mbepsilon_\mbz), \mbepsilon_1,...,\mbepsilon_T \nonumber \\
t_i &= h_i(\mbz) \nonumber \\
y &= g(\mbepsilon_y, \mbz, h_1(\mbz), ..., h_T(\mbz)).
\end{align*}
Both of these models satisfy the independence of treatments
given the shared confounder and have the same joint distribution
on $\mbt, y$. But the second model differs in key way.
It assumes all of the treatments
are due to confounding. Mixtures of these models also have the same
distribution. As the mixing portion of the second model goes to one,
the estimated causal effects would have high variance. Many
\gls{AMI} regularization values have the same prediction error.
To reduce variance, we select the largest \gls{AMI} regularization
value in the class that predicts the best.

\paragraph{Negative Entropy Lower Bound}
\begin{align*}
-\Hb_\mbtheta(\mbz \g \mbt_{-i}) &= \int p(\mbz, \mbt_{-i}) \log p(\mbz \g \mbt_{-i}) d\mbz d\mbt_{-i} \\
&= \int p(\mbz \g \mbt) p(\mbt) \log p(\mbz \g \mbt_{-i}) d\mbz d\mbt
\\
&= \int p(\mbz \g \mbt) p(\mbt) \log \left[\int p(\mbz \g \mbt_{-i}, \hat{t}_i) p(t_i = \hat{t}_i \g \mbt_{-i})d
\hat{t}_i \right] d\mbz d\mbt
\\
&= \int p(\mbz \g \mbt) p(\mbt) \log \left[\int p(\mbz \g \mbt_{-i}, \hat{t}_i) \frac{p(t_i = \hat{t}_i) p(\mbt_{-i} \g t_i = \hat{t}_i)} {p(\mbt_{-i})}
d \hat{t}_i \right] d\mbz d\mbt
\\
&= \int p(\mbz \g \mbt) p(\mbt) \log \left[\int p(t_i = \hat{t}_i) \frac{p(\mbz \g \mbt_{-i}, \hat{t}_i) p(\mbt_{-i} \g t_i = \hat{t}_i)} {p(\mbt_{-i})}
d \hat{t}_i \right] d\mbz d\mbt
\\
&\geq \int p(\mbz \g \mbt) p(\mbt) \left[\int p(t_i = \hat{t}_i) \log \frac{p(\mbz \g \mbt_{-i}, \hat{t}_i) p(\mbt_{-i} \g t_i = \hat{t}_i)} {p(\mbt_{-i})}
d \hat{t}_i \right] d\mbz d\mbt
\\
&= \int p(\mbz \g \mbt) p(\mbt)  p(t_i = \hat{t}_i) \left(\log p(\mbz \g \mbt_{-i}, \hat{t}_i) + \log \frac{p(\mbt_{-i} \g t_i = \hat{t}_i)} {p(\mbt_{-i})}\right)
d \hat{t}_i  d\mbz d\mbt
\\
&= \int p(\mbz \g \mbt) p(\mbt)  p(t_i = \hat{t}_i) \log p(\mbz \g \mbt_{-i}, \hat{t}_i) d \hat{t}_i  d\mbz d\mbt + \int p(\mbz \g \mbt) p(\mbt)  p(t_i = \hat{t}_i) \log \frac{p(\mbt_{-i} \g t_i = \hat{t}_i)} {p(\mbt_{-i})}
d \hat{t}_i  d\mbz d\mbt
\\
& = \int p(\mbz \g \mbt) p(\mbt)  p(t_i = \hat{t}_i) \log p(\mbz \g \mbt_{-i}, \hat{t}_i) d \hat{t}_i  d\mbz d\mbt + C
\end{align*}
The above bound does not require any extra parameters. It may however be loose.
With an auxiliary parameter, we can create a bound that gets tighter as
the auxiliary parameter is optimized. Let $\mbxi_i$ be an auxiliary
parameter and $r$ a distribution parametrized by  $\mbxi_i$, then we have the following lower bound on the negative mutual information
\begin{align*}
-\I(t_i;\mbz \g \mbt_{-i}) &= -\textrm{KL}(p(t_i;\mbz \g \mbt_{-i}) || p(t_i \g
\mbt_{-i}) p(\mbz \g \mbt_{-i})) = -\E_{p(\mbt) p( \mbz \g \mbt)} \log \frac{p(t_i, \mbz \g \mbt_{-i})}{p(t_i \g \mbt_{-i}) p(\mbz \g \mbt_{-i})}
\\
&= -\left(\E_{p(\mbt) p( \mbz \g \mbt)} \log \frac{p(t_i, \mbz \g \mbt_{-i})}{p(t_i \g \mbt_{-i}) p(\mbz \g \mbt_{-i})} + \E_{p(\mbt_{-i})} [\textrm{KL}(p(\mbz \g \mbt_{-i}) || p(\mbz \g \mbt_{-i}))] \right)
\\
&\geq -\left(\E_{p(\mbt) p( \mbz \g \mbt)} \log \frac{p(t_i, \mbz \g \mbt_{-i})}{p(t_i \g \mbt_{-i}) p(\mbz \g \mbt_{-i})} + \E_{p(\mbt_{-i})} [\textrm{KL}(p(\mbz \g \mbt_{-i}) || r(\mbz \g \mbt_{-i} ; \mbxi_i))] \right)
\\
&= -\left(\E_{p(\mbt) p( \mbz \g \mbt)} \log \frac{p(t_i, \mbz \g \mbt_{-i})}{p(t_i \g \mbt_{-i}) p(\mbz \g \mbt_{-i})} + \E_{p(\mbt_{-i}) p(\mbz \g \mbt_{-i})} \log \frac{p(\mbz \g \mbt_{-i})}{r(\mbz \g \mbt_{-i} ; \mbxi_i)} \right)
\\
&= -\left(\E_{p(\mbt) p( \mbz \g \mbt)} \log \frac{p(t_i, \mbz \g \mbt_{-i})}{p(t_i \g \mbt_{-i}) p(\mbz \g \mbt_{-i})} + \E_{p(\mbt) p(\mbz \g \mbt)} \log \frac{p(\mbz \g \mbt_{-i})}{r(\mbz \g \mbt_{-i} ; \mbxi_i)} \right)
\\
&= -\E_{p(\mbt) p( \mbz \g \mbt)} \log \frac{p(t_i, \mbz \g \mbt_{-i})}{p(t_i \g \mbt_{-i}) r(\mbz \g \mbt_{-i} ; \mbxi_i)}
\\
&= -\E_{p(\mbt) p( \mbz \g \mbt)} \log \frac{p(\mbz \g \mbt)}{r(\mbz \g \mbt_{-i} ; \mbxi_i)}
\end{align*}
This is a lower bound because KL divergence is nonnegative.
Maximizing this
bound with respect to $\mbxi_i$ increases the tightness of the bound by
minimizing the KL-divergence. The confounder parameters and
the bound parameters can be simultaneously maximized. The bound is
tight when $r(\mbz \g \mbt_{-i} ; \mbxi_i) = p(\mbz \g \mbt_{-i})$,
so if $r$ is rich enough to contain $p(\mbz \g \mbt_{-i})$. The gap
will be zero.  The introduction of the auxiliary distribution, $r$, is similar to those used in variational inference
\citep{agakov2004auxiliary,salimans2015markov,ranganath2016hierarchical, maaloe2016auxiliary}.

\paragraph{Proposition 1.}
Independence given the confounder means that $t_i$ is independent of $t_j$
given the unobserved confounder. Shared confounding means there is only
a single confounder $\mbz$. Since the form of $f$ is arbitrary, the distribution
on $\mbz$ is arbitrary. Also, since $h_i$ is arbitrary the distribution of $t_j$
given $\mbz$ is arbitrary.
Thus the generative process in \Cref{eq:main-model}
constructs treatments that are conditionally independent given the
confounder.
It can represent any distribution for each treatment
given the confounder. The confounder can also can take any distribution.
This means that \Cref{eq:main-model} can represent any distribution of
treatments that satisfy
both assumptions, of shared confounding and of independence given
confounding.
The outcome function $g$ is arbitrary and so can be chosen
to match any true outcome model.

\paragraph{Gradients of the \gls{MCLBO}.}
The terms in
the \gls{MCLBO} are
all integrals with respect to the distribution $p_\mbtheta(\mbz \g \mbt)$.
To compute stochastic gradients, we differentiate under the
integral sign as in variational inference. For simplicity, we assume that a sample
from $p_\mbtheta(\mbz \g \mbt)$ can be
generated by transforming parameter-free noise $\mbdelta \sim s$ through a
function $\mbz = \mbz(\mbdelta, \mbtheta, \mbt)$. This assumption leads
to simpler gradient computation~\citep{Kingma:2014,rezende2014stochastic}.
The gradient with
respect to $\mbtheta$ can be written as
\begin{align}
\nabla_\mbtheta \cL =& \E_{p(\mbt)} \E_{s(\mbdelta)}\left[\nabla_\mbtheta \mbz
(\mbdelta, \mbtheta, \mbt) \nabla_\mbz \sum_{i=1}^T \log p_\mbbeta(t_i \g \mbz)
\right]
\nonumber \\
&+  \alpha \sum_{i=1}^T \E_{p(\mbt)} \E_{s(\mbdelta)} \left[\nabla_\mbtheta
\mbz(\mbdelta, \mbtheta, \mbt) \nabla_\mbz \log r_{\mbxi_i}(\mbz \g \mbt_{-i})
\right]
\nonumber \\
&- \alpha T \E_{p(\mbt)} \E_{s(\mbdelta)} \left[\nabla_\mbtheta \mbz
(\mbdelta, \mbtheta, \mbt) \nabla_\mbz \log p_\mbtheta(\mbz \g \mbt) \right].
\label{eq:theta-gradient}
\end{align}
Sampling from the various expectations gives a noisy unbiased estimate of the gradient.
The gradient for $\mbbeta$ is much simpler, as the sampled distributions do not depend
on $\mbbeta$:
\begin{align}
\nabla_\mbbeta \cL = \E_{p(\mbt)} \E_{p_\mbtheta(\mbz \g \mbt)} \left[\sum_{i=1}^T \nabla_\mbbeta \log p_\mbbeta(\mbt \g \mbz) \right].
\label{eq:beta-gradient}
\end{align}
Sampling from the observed data then sampling the confounder estimate gives
an unbiased estimate of this gradient. The gradient for $\mbxi_i$ follows
similarly
\begin{align}
\nabla_\mbxi \cL = \alpha \E_{p(\mbt)} \E_{p_\mbtheta(\mbz \g \mbt)} \left[\sum_
{i=1}^T \nabla_{\mbxi_i} [\log r_{\mbxi_i}(\mbz \g \mbt_{-i})] \right]
\label{eq:xi-gradient}
\end{align}

The confounder estimation for a
fixed value of $\alpha$ is summarized in Algorithm \ref{alg:confound-lower-bound}.

\paragraph{Equivalent Confounders.}
Invertible transformations of a random variable preserve the
information in that random variable. Take two
distributions for computing the stochastic confounder
$\mbz_1 \sim p_1(\cdot \g \mbt)$ and $\mbz_2 \sim p_2(\cdot \g \mbt)$
where $\mbz_2$ can be written as an invertible
function of $\mbz_1$. These two distributions have
equivalent information for downstream tasks,
such as building the
outcome model or conditioning on the confounder.
This equivalence means we have choice on which member
in the equivalence class we choose.
One way to narrow the choice
is to enforce that the dimensions of $\mbz$ are
independent by minimizing total correlation.

\paragraph{Connection to Factor Analysis.}
Factor analysis methods work by specifying a generative model
for observations that independently generate each dimension of each
observation. In its most general form this model is
\begin{align*}
\mbz_n = f(\mbepsilon_z), \\
\mbt_{n, i} = h_i(\mbepsilon_y, \mbz_n).
\end{align*}
Inference in this model matches the reconstruction term inside
our confounder estimator with a $KL$-divergence regularizer. If
we allow for the parameters of the prior on $\mbz$ to be learned
to maximize the overall likelihood, and if
$\mbz$'s dimensions are independent, then inference corresponds
to minimizing the reconstruction \cref{eq:mc} with
a total correlation style penalty.

There are many ways to choose the complexity of the factor model.
One choice is to find the smallest complexity model
that still gives good predictions of $\mbt_{i}$ given $\mbt_{-i}$
(like document completion evaluation in topic models \citep{wallach2009evaluation}).
Here complexity is measured in terms of
the dimensionality of $\mbz$ and the complexity of $h_i$ and $f$.
This choice tries to minimize the amount of information retained
in $\mbz$, while still reconstructing the treatments well. This
way to select the factor analysis model's complexity is a type of
\gls{AMI} regularization. However, selecting
discrete parameters like dimensionality give less fine-grained control
over the information rates.

\paragraph{Proposition 2.}
If the data are conditionally i.i.d., then in the true model
$\mbz$ concentrates as the number of treatments
goes to infinity. In this setting, we can learn the model from Proposition 1 using the \gls{MCLBO}.
This follows because the information each treatment provides goes to zero as $T \to \infty$
since they are conditionally i.i.d., thus
the true confounder (and posterior), up to information equivalences,
is simply a point that maximizes the reconstruction term in the \gls{MCLBO}
subject to asymptotically zero \gls{AMI}. Identifying the parameters
of the confounder estimator requires that $N \to \infty$.
This shows
outcome estimation corresponds to simple regression with treatments
and confounder (up to an information equivalence), which correctly estimates the causal effects as $N \to \infty$.

\paragraph{Estimating $\mbepsilon_i$.}
The $\mbepsilon_i$ estimation requires finding parameters $\mblambda$ and $\mbnu$ that maximize
\begin{align*}
\E_{p(\mbt) p_\mbtheta(\mbz \g \mbt) \prod_i p_\mblambda(\mbepsilon_i \g \mbz, \mbt_i)} \left[\sum_{i=1}^T \log p_\mbnu(t_i \g \mbz, \mbepsilon_i) \right], \textrm{such that } \I[\mbepsilon_i, \mbz] = 0.
\end{align*}
The constraint can be baked into a Lagrangian with parameter $\kappa$,
\begin{align*}
\E_{p(\mbt) p_\mbtheta(\mbz \g \mbt) \prod_i p_\mblambda(\mbepsilon_i \g \mbz, \mbt_i)} \left[\sum_{i=1}^T \log p_\mbnu(t_i \g \mbz, \mbepsilon_i) \right] - \kappa \I[\mbepsilon_i, \mbz].
\end{align*}
The mutual information can be split into entropy terms:
\begin{align*}
\I[\mbepsilon_i, \mbz] = \Hb(\mbepsilon_i) - \Hb(\mbepsilon_i \g \mbz).
\end{align*}
We can use the entropy bounds with auxiliary
distributions on the conditioning set.
These bounds work with a distribution over the reverse conditioning set in this
case $r(\mbt \g \mbz, \mbepsilon_i)$. For this, we can use the reconstruction distribution
$p_\mbnu(t_i \g \mbz, \mbepsilon_i)$ and the fact that $p(\mbz)$ and $p(\mbt)$ do not
depend on the parameters $\mbnu$ and $\mbtheta$.

\paragraph{Confounder Parameterization and Simulation Hyperparameters.}
We limit the confounder to have similar complexity as \gls{PCA}.
We do this by using a confounder distribution with normal noise,
where we restrict the mean of the confounder estimate to be a linear function of the treatments $\mbt$. The variance
is independent and controlled by a two-layer (for second moments) neural
network. We similarly limit the likelihoods and outcome model to have
three-layer means and fixed variance.

For the remaining simulation hyperparameters, we set $W$ and $b$ to be the absolute
value of draws from the standard normal. The weights $b_\mbepsilon$ is scaled by
a decreasing
sequence of $t^{-0.6}$ to ensure finite variance. We fix the
simulation standard deviation to
$0.02$ and fix outcome standard deviation to $0.1$.

\end{document}